%% file: main.tex
\documentclass[journal]{IEEEtran}
\ifCLASSOPTIONcompsoc
\usepackage[nocompress]{cite}
\else
\usepackage{cite}
\fi
\ifCLASSINFOpdf
\else
\fi
\usepackage{multirow,palatino, epsfig, latexsym, algorithmic, algorithm , epstopdf, amsmath, amssymb, color, amsthm,graphicx}

\newcommand{\ignore}[1]{}

\newtheorem{myDefinition}{Definition}

\begin{document}
\title{Linear centralization classifier}
	
\author{Mohammad Reza Bonyadi, Viktor Vegh, David C. Reutens
		
\thanks{All authors are with the Centre for Advanced Imaging (CAI), the University of Queensland, Brisbane, QLD 4072, Australia. M. R. Bonyadi (reza@cai.uq.edu.au, rezabny@gmail.com) is also with the Optimisation and Logistics Group, The University of Adelaide, Adelaide 5005, Australia. }}

	
\IEEEtitleabstractindextext{%
\begin{abstract}
A classification algorithm, called the Linear Centralization Classifier (LCC), is introduced. The algorithm seeks to find a transformation that best maps instances from the feature space to a space where they concentrate towards the center of their own classes, while maximimizing the distance between class centers. We formulate the classifier as a quadratic program with quadratic constraints. We then simplify this formulation to a linear program that can be solved effectively using a linear programming solver (e.g., simplex-dual). We extend the formulation for LCC to enable the use of kernel functions for non-linear classification applications. We compare our method with two standard classification methods (support vector machine and linear discriminant analysis) and four state-of-the-art classification methods when they are applied to eight standard classification datasets. Our experimental results show that LCC is able to classify instances more accurately (based on the area under the receiver operating characteristic) in comparison to other tested methods on the chosen datasets. We also report the results for LCC with a particular kernel to solve for synthetic non-linear classification problems.
\end{abstract}
		
}
\maketitle
\IEEEdisplaynontitleabstractindextext
\IEEEpeerreviewmaketitle

\section{Introduction}
\label{sec:intro}
\input{intro_background.tex}
	
\section{Proposed algorithm}
\label{sec:proposed}
\input{proposed.tex}

\section{Experiments and results}
\label{sec:experiment}
\input{experiments.tex}

\section{Conclusion and future work}
\label{sec:conclusions}
We have introduced the linear centralization classifier, LCC, for solving two-class classification problems. Our approach provides a linearized formulation for classification and we benchmarked our method against existing classification approaches. Based on four state-of-the-art classification methods and eight standard classification datasets, we were able to classify instances more accurately as measured using the area under the curve of the receiver operating characteristic. We also introduced a kernel version of LCC that is able to classify non-linearly separable instances using a standard kernel, the radial basis function. This new classification method may potentially be used to improve and speed-up classification of datasets. Future direction involves extending the method to deal with multiclass classification and inclusion of structures of the samples in different classes for further accuracy improvement.

\input{appendix.tex}

\small

\bibliographystyle{IEEEtran}
\bibliography{References}

\end{document}

%% file: intro_background.tex
\IEEEPARstart{T}{he} ultimate goal of a supervised classification algorithm is to identify the class to which each instance belongs based on a given set of correctly labeled instances. A discriminative classifier \cite{ng2002discriminative} is defined as follows:	

\begin{myDefinition}
(Discriminative classifier)
Let $ S,S_{1},S_{2},...,S_{c} $ be sets of instances such that, $ \forall i,j \in \{1,...,c\} $, $ i \ne j $, $ |S_i|=m_i $, $ |S|=m $, $ S_{i} \cap S_{j} = \emptyset $, and $ \cup_{i=1}^{c}S_i=S $. A classifier $\psi_\beta:S \to \{1,...,c\}$ aims to guarantee
\begin{equation*}
	\forall i \in \{1,...,c\}, \forall x \in S_i, P(\psi_\beta(x)=i|x)=1,
\end{equation*}	
\noindent where $\beta$ is a set of configurations for the procedure $\psi_\beta(x)$, and $P$ is the probability measure.
\end{myDefinition}
The classifier $ \psi_\beta(\cdot) $ is usually a combination of an optimization problem, $ \Omega_\beta $, a transformation $ \mathcal{M}_\beta: S \to \hat{S} $, and a discriminator $ \mathcal{D}: \hat{S} \to \{1,...,c\} $. The solution to $ \Omega_\beta $ yields $ \beta $ that can transform any given instance $ x $ to $ \hat{x} $ through $ \hat{x} = \mathcal{M}_\beta(x) $, which finally, can be mapped into the class label by $ \mathcal{D}(\hat{x}) $. In reality, the true class of only a subset of $ S $ is known (the training set). It is hence challenging to find a transformation $\beta$ for which $\psi_\beta(\vec{x})$ is the true class of all $ \vec{x} $ in $ S $, including the ones that are not in the training set (unseen instances). Therefore, the generality of an optimized $ \beta $ depends on the assumptions made to formulate $ \Omega_\beta $ and the given training set itself. Hence, the optimal choice of the classifier is problem dependent, often achieved by a cross validation \cite{kohavi1995study}. Various basic assumptions to formulate $ \Omega_\beta $ have led to many classification methods being proposed such as multi-layer perceptrons \cite{haykin2004comprehensive}, decision trees \cite{safavian1991survey}, support vector machines \cite{suykens1999least}, and extreme learning machines \cite{huang2006extreme}, amongst others.

We consider a special case of classification problems where all members of $ S $ are in $ \mathbb{R}^n $ (so called feature space), and each instance in $ S $ is represented by a vector. We also assume that feasible values for $ x_j $ (called a variable throughout this paper), the $ j^{th} $ element of the instance $ \vec{x} $, are ordered by the operator "$ \le $" (i.e. $ x_j $ is not categorical). We only focus on $ c=2 $ (binary classification, labeled by $-1$ and $1$), that can be extended to multiclass classification by using one-vs-all or one-vs-one strategies \cite{multiclassSVMComparisons}. 

We propose a new classification algorithm that aims to find a transformation that maximizes the distance between the centers of the classes whilst ensuring instances are close to their class center. We define the constraints and objective associated with this classifier as a linear program. As such, we refer to it as the Linear Centralization Classifier (LCC). We validate LCC using eight standard benchmarking binary classification datasets and compare the results with six other classification methods. 

We structure the paper as follows: Section \ref{sec:background} provides a background on the classification methods used for comparison. Section \ref{sec:proposed} details our proposed method. Section \ref{sec:experiment} reports and discusses the results of the comparisons between multiple classification methods based on eight standard benchmark classification problems. Section \ref{sec:conclusions} concludes the paper and points to potential future directions.

\section{Background}
\label{sec:background}
This section provides background information on existing classification methods and associated kernelization.
\subsection{Established classification methods}
We describe in brief popular classification methods used herein for benchmarking and comparison.
		
\subsubsection{Support vector machines (SVM)}
SVM aims to find a hyperplane $ \mathcal{H} $ defined by the normal vector $ \vec{\omega} $ that separates the instances such that the distance between the closest instances (i.e., support vectors) and $ \mathcal{H} $ from each class is maximized \cite{suykens1999least}. The separation is determined by the sign of $ \vec{x}\vec{\omega}^T + r $ ($ T $ is transpose) indicating the side of $ \mathcal{H} $ on which the instance, $ \vec{x} $, belongs. Formally, $ \beta=<\vec{\omega}, r> $, $ \mathcal{M}_\beta(\vec{x}) = \vec{x}\vec{\omega}^T + r $ and $ \mathcal{D}(\vec{x})=sign(\mathcal{M}_\beta(\vec{x})) $. The optimization problem $ \Omega_\beta $ for SVM is defined as:	
\begin{equation}
	\begin{array}{ll}	
	\min_\beta{\lambda||\vec{\omega}||^2+\frac{1}{m}\sum_{i=1}^{m}\epsilon_i}  \\ 
	s.t.~ \forall i, \begin{cases}
	y^{(i)}\mathcal{M}_\beta(\vec{x}^{(i)}) \ge 1-\epsilon_i\\
	\epsilon_i \ge 0
	\end{cases}
	\end{array}.		
\end{equation}
This optimization problem has $m$ inequality constraints, $m$ boundaries, and $m+n+1$ variables, which can be solved effectively using quadratic programming. Note that SVM seeks to maximize between-class margins to ensure correct classification of (ideally) all given instances in the training set, ignoring the distribution by which the instances may have been generated.

\subsubsection{Linear discriminant analysis (LDA)}
The aim of LDA is to find $ \beta $ such that, in the transformed space, the distance between the centers of the classes is maximized while the spread of instances within the class is minimized. Assuming $ p=1 $ and the conditional probabilities $ P(\vec{x}^{(i)}|y^{(i)}=-1) $ and $ P(\vec {x}^{(i)}|y^{(i)}=1) $ ($y^{(i)}$ is the label of $\vec{x}^{(i)}$) are both normally distributed with mean and covariance parameters $ \left({\vec {\mu }}_{1},\Sigma_1\right) $ and $ \left({\vec {\mu }}_{2},\Sigma_2\right) $, Fisher \cite{fisher1936use} proved that $ \beta=(\Sigma_1+\Sigma_2)^{-1}(\vec{\mu}_2-\vec{\mu}_1) $ and $ k={\frac {1}{2}}{\vec {\mu }}_{2}^{T}\Sigma_2^{-1}{\vec {\mu }}_{2}-{\frac {1}{2}}{\vec {\mu }}_{1}^{T}\Sigma_1^{-1}{\vec {\mu }}_{1} $ leads to maximizing of $\frac{W^TS_BW}{W^TS_WW}$, where $S_W=(\Sigma_1+\Sigma_2)$ and $S_B=(\vec{\mu}_2-\vec{\mu}_1)(\vec{\mu}_2-\vec{\mu}_1)^T$. In this case, $ \beta $ is considered to be the norm of a hyperplane that discriminates the two classes and $ k $ shifts the hyperplane to be between the two classes, i.e., $ \vec{x}\beta>k $ if the instance $ \vec{x} $ belongs to class $ 1 $ (i.e., $\Omega_{\beta,k}$ seeks to maximize $\frac{W^TS_BW}{W^TS_WW}$, and $\mathcal{D}(\vec{x})=sign(\mathcal{M}_{\beta,k}(\vec{x}))=\vec{x}\beta-k$). 

If $ \Sigma_1 $ and $ \Sigma_2 $ are small then $ (\Sigma_1+\Sigma_2)^{-1} $ becomes singular and $ (\vec{\mu}_2-\vec{\mu}_1) $ vanishes, i.e. a solution that leads to a singular $ (\Sigma_1+\Sigma_2)^{-1} $ dominates over all other solutions, no matter the distance between the centers of the classes \cite{gao2006direct}. This is undesirable as it is important for class centers to be separated for classes to be distinguishable. Such a scenario occurs particularly when the number of instances in a class is smaller than the number of dimensions $n$. The threshold $ k $ is effective only if the distribution of the instances associated with each class are similar, which may not be the case for certain datasets. To overcome the singularity issue, the covariance matrices are replaced by a regularized term ($ \lambda\Sigma + (1-\lambda\Sigma) $) which ensures the impact of between class separation does not vanish \cite{guo2006regularized}.

\ignore{
\subsubsection{Linear programming-based classifiers}
\cite{zhou2002linear}
\cite{fung2004feature}
}

\subsection{State-of-the-art extensions}
\subsubsection{Structural minimax probability machine (SMP)}
On the one hand, minimax probability machine (MPM) is a type of discriminative classifier that aims to minimize the maximum misclassification probability of instances \cite{lanckriet2002robust}. Unlike SVM, and similar to LDA, MPM attempts to find a generalizable margin by paying attention to the distributions within classes rather than the instances themselves. On the other hand, there exists evidence \cite{xue2011structural} suggesting that the structure of the instances in different classes provides important information for the design of generalizable transformations for classification. Structural minimax probability machine (SMP) \cite{gu2017structural} makes use of the structural information, approximated by two finite mixture models in each class, in the context of MPM for classification of instances. This idea has shown to be very effective on a set of standard datasets. 

\subsubsection{Twin SVM}
Different from SVM, Twin SVM (TSV) \cite{shao2011improvements,khemchandani2007twin} seeks a pair of hyperplanes, $\mathcal{H}_{-1}$ and $\mathcal{H}_1$, such that $\mathcal{H}_{-1}$ is closer to the instances from class -1 in comparison to the instances from class 1, while $\mathcal{H}_{1}$ is closer to the instances from class 1 comparing to the instances from class -1. The discriminator then is used to calculate which hyperplane is closer to an instance. A recent extension \cite{shao2015weighted} of TSV is based on introducing a weighted linear loss to the formulation of TSV (hence, called WSV throughout the paper) instead of the Hinge loss, reducing the quadratic problem to a linear one.  
	
\subsubsection{Discriminative elastic net (DEL)}
Elastic net incorporates the least absolute shrinkage and selection operator (LASSO) \cite{tibshirani1996regression} and Tikhonov regularization \cite{bishop2006pattern} terms into the linear (or logistic) regression formulation \cite{walker1967estimation} for regression or classification problems. For classification purposes, elastic net usually performs optimization in relation to the true class labels, restricting the algorithm to binary targets. This restriction was resolved in \cite{zhang2017discriminative} where a term was used to relax the class labels. The optimization problem associated with this classification process was then introduced and solved via an iterative procedure. 

\subsection{Kernelization}
\label{sec:kernalization}
The use of kernels plays an important role in classifying instances that are not linearly separable \cite{smola1998learning}. For such problems, a transformation $\phi: \mathbb{R}^n \to \mathbb{R}^p$ is defined in a way that the instances transformed by $\phi$ are linearly separable. If $p \gg n$ then the classification task would involve larger computational complexity. To avoid such increase in the computational complexity, it is beneficial to compute the higher dimensional space inner product without mapping instances by $\phi$. Hence, one seeks to define a 'kernel', a transformation $\kappa:\mathbb{R}^n\times \mathbb{R}^n \to \mathbb{R}$, $\kappa(\vec{x}^{(i)},\vec{x}^{(j)})=\phi(\vec{x}^{(i)})\cdot\phi(\vec{x}^{(j)})$, that performs the inner product in the the $ p $ dimensional space implicitly. The challenge then is to find an expression for $\kappa$ which does not require the computation of $\phi(\vec{x}^{(i)})$ and $\phi(\vec{x}^{(j)})$. In view of some restrictions \cite{mercer1909functions}, a number of kernels have been developed for various classification applications all of which aim to reduce the computational complexity. For example, the Radial Basis Function (RBF) kernel has the form:
\begin{equation}
    \mathcal{K}(\vec{x}^{(i)},\vec{x}^{(j)}) = e^{\frac{-||\vec{x}^{(i)}-\vec{x}^{(j)}||^2}{2\sigma^2}},
\end{equation}
\noindent where $\sigma$ controls how $\kappa$ changes with changes in distance between $\vec{x}^{(i)}$ and $\vec{x}^{(j)}$. 

%% file: proposed.tex
In this section we describe $\Omega_\beta$, $ \mathcal{M}_\beta $, and $ \mathcal{D}$ for our proposed classifier.

\subsection{Classification by centralizing instances}
The aim of our proposed method is to optimize transformation parameters of $ \mathcal{M}_\beta $ such that the transformed instances appear closer to their class center while the distance between the class centers in the transformed space is maximized. This is a combination of the SVM and LDA aims. Similar to LDA, we maximize the distance between the class centers. Unlike LDA and similar to SVM, we minimize the penalty associated with the total number of misclassified instances. This approach reduces the norm of the covariance matrix, whilst ensuring instance misclassification is minimized. Having found an optimal transformation, $ \mathcal{M}_\beta $ transforms instances to a space where they are more "centralized" to class centers (hence, the name, centralization classifier). Any given instance $ \vec{x}^{(0)} $ can then be assigned to one of the classes using the discriminator $ \mathcal{D}$ that is simply the distance between the given instance and the centers of the two classes ($ \vec{C}_{-1} $ and $ \vec{C}_{1} $) in the transformed space. 

We define the optimization problem associated with the centralization classifier ($\Omega_\beta$) as:
\begin{equation}
\label{Eq:optimizationProblem}
	\begin{array}{ll}	
		\min\limits_{\beta}{-\left(dis(\hat{C}_{-1},\hat{C}_{1})\right)} \nonumber \\ 
		s.t.~ \forall i, \begin{cases}
		dis(\hat{x}^{(i)},\hat{C}_{-1}) < dis(\hat{x}^{(i)},\hat{C}_{1}) & y^{(i)}=-1\\
		dis(\hat{x}^{(i)},\hat{C}_{1}) < dis(\hat{x}^{(i)},\hat{C}_{-1}) & y^{(i)}=1
		\end{cases}
	\end{array}	
\end{equation}
\noindent where $ \hat{z} $ is $ \mathcal{M}_\beta(\vec{z}) $. Also, $ dist(a,b) $ provides a distance measure ($ dist: \mathbb{R}^n\times\mathbb{R}^n \to \mathbb{R}$), e.g., the Euclidean distance between $ \vec{a} $ and $ \vec{b} $, $ ||\vec{a}-\vec{b}|| = \sqrt{(\vec{a}-\vec{b})^T(\vec{a}-\vec{b})} $. 

Consider a linear transformation, $ \hat{z}=\mathcal{M}_\beta(\vec{z})=\vec{z}\beta $, where $ \beta $ is a $ n \times 1 $ transformation operator and $ \vec{z} $ is $ 1 \times n $, then it is possible to rewrite Eq. \ref{Eq:optimizationProblem} as:	
\begin{equation}
	\label{Eq:hardMarginFqcc}
		\begin{array}{ll}	
		\min\limits_{\beta}{\left(-||\hat{C}_{-1}-\hat{C}_{1}||\right)} \\ 
		s.t. ~ \forall i, ~ y^{(i)}\left(||\hat{x}^{(i)}-\hat{C}_{1}|| - ||\hat{x}^{(i)}-\hat{C}_{-1}||\right) < 0
		\end{array}		
\end{equation}
This optimization problem involves $ m $ constraints and $ n $ variables for the classification of instances. An optimized $ \beta $ ensures that the transformed centers, $ \vec{C}_{-1}\beta $ and $ \vec{C}_{1}\beta $, are as far away as possible from each other while the distance between $ \vec{x}^{(i)}\beta $ and the center to which $ \vec{x}^{(i)} $ belongs is smaller than the distance between $ \vec{x}^{(i)}\beta $ and the other center. As such, instances can be distinguished by (see Section \ref{sec:alternativediscriminators} for other alternatives):	
\begin{equation}
\label{Eq:distinguisher}
y^{(0)}=\begin{cases}
-1 & ||\hat{x}^{(0)}-\hat{C}_{-1}|| \le ||\hat{x}^{(0)}-\hat{C}_{1}||\\
1 & otherwise\\
\end{cases}
\end{equation}
To guarantee existence of at least one feasible solution, a necessary condition for Eq. \ref{Eq:hardMarginFqcc} is $ y^{(i)}\left(||\hat{x}^{(i)}-\hat{C}_{1}|| - ||\hat{x}^{(i)}-\hat{C}_{-1}||\right) < 0 $ for all $ i $, which might not be achievable in real-world classification problems. Hence, we define a "slack" variable and incorporate its values to the objective function of Eq. \ref{Eq:hardMarginFqcc} as:	
	
\begin{equation}
	\label{Eq:softMarginFqcc}
	\begin{array}{ll}	
	\min\limits_{\beta,\epsilon_1,...,\epsilon_m}{\left(-||\hat{C}_{-1}-\hat{C}_{1}||\right)}+\lambda\sum_{i=1}^{m}\epsilon_i \\ 
	s.t. ~ \forall i,\begin{cases}
	y^{(i)}\left(||\hat{l}_{1}^{(i)}|| - ||\hat{l}_{-1}^{(i)}||\right) \le \epsilon_i \\
	\epsilon_i \ge \sigma
	\end{cases}	
	\end{array}		
\end{equation}	
\noindent where $ \hat{l}_1^{(i)}=\vec{x}^{(i)}\beta-\vec{C}_{1}\beta $, $ \hat{l}_{-1}^{(i)}=\vec{x}^{(i)}\beta-\vec{C}_{-1}\beta $, $ \hat{C}_k=\vec{C}_k\beta $, $ \sigma<0 $ is close to zero, and $ \lambda>0 $. 
	
The boundaries for $ \beta $ can be set to $ [-1, 1]^n $ without loss of generality. The reason is that $ \beta $ does not shift the instances (intercept equal to zero), hence, for any $\beta_0$, there exists $\beta=\frac{\beta_0}{||\beta_0||}$ that acts the same as $\beta_0$ in satisfying the constraints and optimizing the objective value of Eq. \ref{Eq:softMarginFqcc}. Thereby, LCC can be forumated as:	
\begin{equation}
	\label{Eq:finalFqcc}
	\begin{array}{ll}	
	\min\limits_{\beta,\epsilon_1,...,\epsilon_m}{\left(-||\hat{C}_{-1}-\hat{C}_{1}||\right)}+\lambda\sum_{i=1}^{m}\epsilon_i \\ 
	s.t. ~ \forall i,\begin{cases}
	y^{(i)}\left(||\hat{l}_{1}^{(i)}|| - ||\hat{l}_{-1}^{(i)}||\right) \le \epsilon_i \\
	\epsilon_i \ge \sigma \\
	\beta \in [-1, 1]^n
	\end{cases}	
	\end{array}		
\end{equation}
This optimization problem, together with the discriminator $\mathcal{D}$ introduced in Eq. \ref{Eq:distinguisher} and considering $\mathcal{M}_\beta(\vec{k})=\vec{k}\beta$, defines our proposed classifier. We refer to this as the Fully Quadratic Centralization Classifier (FQCC). We further simplify FQCC in Section \ref{sec:linearizingConstraints}.
	
\subsection{Role of parameters}
If an instance $ \vec{x}^{(i)} $ in the transformed space is closer to its class center than to the other center by at least $ |\sigma| $ then $ \epsilon_i= \sigma $. When this is true for all $ i $ then $ \lambda\sum_{i=1}^{m}\epsilon_i = \lambda m \sigma $, a constant. The optimum solution is therefore obtained when the distance between class centers is maximized. If $ \sigma < \epsilon_i < 0 $ then $ \vec{x}^{(i)} $ has been classified correctly, however it is on the margin between the two classes. That is, in the transformed space $ \vec{x}^{(i)} $ is closer to its correct center than its incorrect center by a value smaller than $ |\sigma| $. In the case of $ \epsilon_i > 0 $ then $ \vec{x}^{(i)} $ is closer to the incorrect center. In the both latter cases the objective is penalized. Hence, $ |\sigma| $ formalizes the margin between two classes in the transformed space and a larger value for $ |\sigma| $ leads to a larger margin between the classes. The impact of misclassification of an instance can be calculated by $ \lambda(\epsilon_i-\sigma) $. Hence, $ \text{argmax}_i\{\lambda(\epsilon_i-\sigma)\} $ yields the index of the instance that has the largest misclassification impact on the objective function and, could be treated as an outlayer.

A large $ |\sigma| $ may lead to misclassification of instances or, in extreme cases, an infeasible linear problem. Thus, $ \sigma $ needs to be set for each application. An obvious upper bound for $ |\sigma| $ is $ ||\vec{C}_{-1}-\vec{C}_{1}|| $, since larger values for $ |\sigma| $ lead to the undesirable outcome of misclassification of all instances. 

The role of $ \lambda $ is to control the balance between satisfying the constraints and maximizing the distance between the centers. A larger $ \lambda $ places a larger emphasis on classifying instances correctly. This may not, however, always be a good choice, as many instances may be outlayers and eforcement of constraints could lead to suboptimal transformations. 
	
\subsection{Linearization of objective and constraints}
\label{sec:linearizingConstraints}
Only a locally optimal solution can be found for Eq. \ref{Eq:finalFqcc} using the interior-point method \cite{wright1997primal}, since positive definiteness of the objective and constraints cannot be guaranteed. We therefore linearize all of the FQCC constraints and the objective to overcome this problem.
	
Assuming $ y^{(i)} = 1 $, the main aim of the constraint $ ||(\vec{x}^{(i)}-\vec{C}_{1})\beta|| < ||(\vec{x}^{(i)}-\vec{C}_{-1})\beta|| $ in Eq. \ref{Eq:hardMarginFqcc} is to ensure that $ \hat{x}^{(i)} $ is closer to $ \hat{C}_{1} $ than $ \hat{C}_{-1} $. It is hence possible to replace $ ||\hat{x}^{(i)}-\hat{C}_{-1}|| < ||\hat{x}^{(i)}-\hat{C}_{1}|| $ by $ \hat{x}^{(i)}-\frac{\hat{C}_{-1}+\hat{C}_{1}}{2} < 0 $ while ensuring $ \hat{C}_{-1}<\hat{C}_{1} $\footnote{Let $ a,b,c \in \mathbb{R} $ and $ b<c $. Because $ a $, $ b $, and $ c $ are scalars then the inequality $ ||a-b||<||a-c|| $ is  reduced to $ |a-b|<|a-c| $. Because $b<c$, this inequality holds iff $ a $ is closer to $ b $ than $ c $, that is $ a<\frac{b+c}{2} $. Hence, $ ||a-b||<||a-c|| $ iff $ a-\frac{b+c}{2}<0 $.}. Following a similar rationale, one can also show that $ \hat{x}^{(i)}-\frac{\hat{C}_{-1}+\hat{C}_{1}}{2} > 0 $ iff $ ||\hat{x}^{(i)}-\hat{C}_{-1}|| > ||\hat{x}^{(i)}-\hat{C}_{1}|| $. This means that we can rewrite the optimization problem in Eq. \ref{Eq:hardMarginFqcc} as:
\begin{equation}
	\label{Eq:QCC}
	\begin{array}{ll}	
	\min\limits_{\beta}{\left(-||\vec{C}_{-1}\beta-\vec{C}_{1}\beta||\right)} \\ 
	s.t.~ \begin{cases}
	\forall i, ~ y^{(i)}\left(\vec{x}^{(i)}\beta-\vec{l}\beta\right) > 0 \\
	\vec{C}_{-1}\beta < \vec{C}_1\beta \\
	\beta \in [-1, 1]^n	
	\end{cases}	
	\end{array}		
\end{equation}
\noindent where $ \vec{l}=\frac{\vec{C}_{-1}+\vec{C}_{1}}{2} $. Here, constraints are linear and the solution is also the solution to the problem defined in Eq. \ref{Eq:hardMarginFqcc}. 

Now let us assume that the best solution for Eq. \ref{Eq:finalFqcc} is $ \beta_0 $, and $ \beta_0 $ violates $ \vec{C}_{-1}\beta < \vec{C}_1\beta $. Then, $ -\beta_0 $ acts the same way as $ \beta_0 $, which is to satisfy all constraints in Eq. \ref{Eq:finalFqcc} and maximize the distance between centers while satisfying $ \vec{C}_{-1}\beta < \vec{C}_1\beta $ as well. Thereby, the constraint $ \vec{C}_{-1}\beta < \vec{C}_1\beta $ does not impact the generality of Eq. \ref{Eq:QCC}.
	
Since $ \vec{C}_{-1}\beta < \vec{C}_1\beta $ is a guaranteed constraint, $ ||\vec{C}_{-1}\beta-\vec{C}_{1}\beta|| $ is maximized iff $ \vec{C}_{1}\beta-\vec{C}_{-1}\beta $ is maximized. Hence, the optimization problem for FQCC can be simplified further:
\begin{equation}
	\begin{array}{ll}	
	\min\limits_{\beta}{\left(\vec{C}_{-1}\beta-\vec{C}_{1}\beta\right)}\nonumber \\ 
	s.t. ~ \begin{cases}
	\forall i, ~ y^{(i)}\left(\vec{x}^{(i)}\beta-\vec{l}\beta\right) > 0 \\
	\vec{C}_{-1}\beta < \vec{C}_1\beta\\
	\beta \in [-1, 1]^n
	\end{cases}
	\end{array}		
\end{equation}
Intuitively, the optimal solution to this linear program ensures that 
\begin{itemize}
	\item $ \vec{C}_{-1} $ is smaller than $ \vec{C}_{1} $ (constraint $ \vec{C}_{-1}\beta < \vec{C}_1\beta $).
	\item $ \vec{x}^{(i)} $ are closer to their own centers (constraints $ y^{(i)}\left(\vec{x}^{(i)}\beta-l\beta\right) > 0 $).
	\item The distance between the centers is maximized. 
\end{itemize}
As $ y^{(i)}\left(\vec{x}^{(i)}\beta-l\beta\right) > 0 $ may not all be satisfiable, we introduce $ \epsilon_i $:
\begin{equation}
	\label{Eq:LCC}
	\begin{array}{ll}	
	\min\limits_{\beta,\epsilon_1,...,\epsilon_m}{\left(\hat{C}_{-1}-\hat{C}_{1} + \lambda\sum_{i=1}^{m}\epsilon_i\right)} \\
	s.t. ~ \begin{cases}
	\forall i, ~ y^{(i)}\left(\hat{l}-\hat{x}^{(i)}\right) \le \epsilon_i \\
	\forall i, ~ \epsilon_i \ge \sigma \\
	\hat{C}_{-1} \le \hat{C}_1 + \sigma\\
	\beta \in [-1, 1]^n\\
	\end{cases} 
	\end{array}		
\end{equation}
\noindent where $ \hat{z}=\vec{z}\beta $. This equation is completely linear and can optimally be solved using any linear programming method (e.g., simplex-dual).

While the discriminator introduced in Eq. \ref{Eq:distinguisher} is effective in distinguishing instances transformed by $ \beta $ optimized by Eq. \ref{Eq:LCC}, it can also be linearized: 
\begin{equation}
\label{Eq:distinguisherLCC}
y^{(0)}=\begin{cases}
-1 & \hat{x}^{(0)}<\frac{\hat{C}_{-1}+\hat{C}_{1}}{2} \\
1 & otherwise\\
\end{cases}
\end{equation}

The combination of the optimization problem $\Omega_\beta$ in Eq. \ref{Eq:LCC}, the discriminator $\mathcal{D}$ in Eq. \ref{Eq:distinguisherLCC}, and a linear transformation $\mathcal{M}_\beta$ define the Linear Centralization Classifier (LCC). The optimization problem associated with LCC is a linear program, and Eq. \ref{Eq:LCC} involves $ m+1 $ inequality constraints, $ m+n $ boundaries, and $ m+n $ variables.

\subsection{A visual overview through an example}
\begin{figure}
	\centering
	\includegraphics[width=0.49\textwidth]{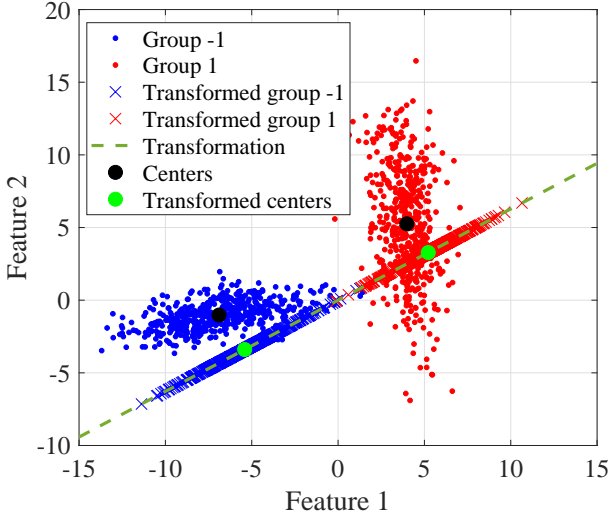} 
	\caption{After the transformation, the instances are mapped into a line, represented by $ \beta $.}
	\label{fig:LCC-example-real}
\end{figure}

We generate a synthetic dataset for a binary classification problem for the purpose of demonstrating how LCC works. We use a multivariate normal distribution,\footnote{The covariance matrices and the means have been selected by some trial to illustrate the procedure of LCC as clear as possible and they do not have any other specific characteristics.} with covariance matrix of $ \begin{bmatrix}
	0.94&0.34\\
	-0.34&3.76\\
	\end{bmatrix}$ and $ \mu=<4,5> $ for group $ -1 $, and with covariance matrix 
	$ \begin{bmatrix}
	-2.57&-0.77\\
	0.767&-0.64\\
	\end{bmatrix} $ and $ \mu=<-7,-1> $ for group 1, and characteristics are shown in Figure \ref{fig:LCC-example-real}. 
We then solve the LCC optimization problem (i.e., Eq. \ref{Eq:LCC}) to find $ \beta $, the optimal transformation. Note, $ \beta $ transforms instances from an $ n $-dimensional space to a one-dimensional space where the instances are separable by the discriminator $ \mathcal{D} $ in Eq. \ref{Eq:distinguisherLCC}. 

\begin{figure}
	\begin{tabular}{c}
	\includegraphics[width=0.45\textwidth]{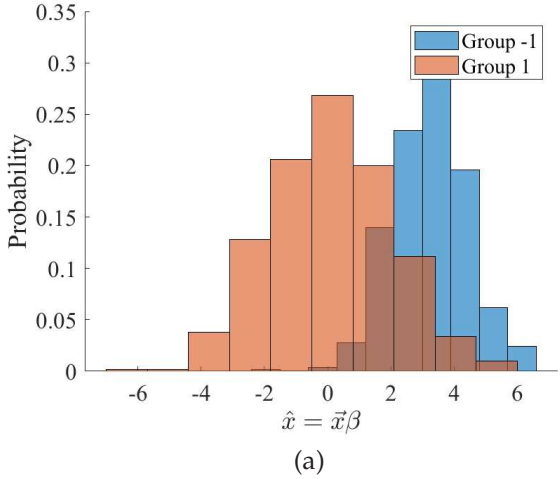}\\
	(a)\\		
	\includegraphics[width=0.45\textwidth]{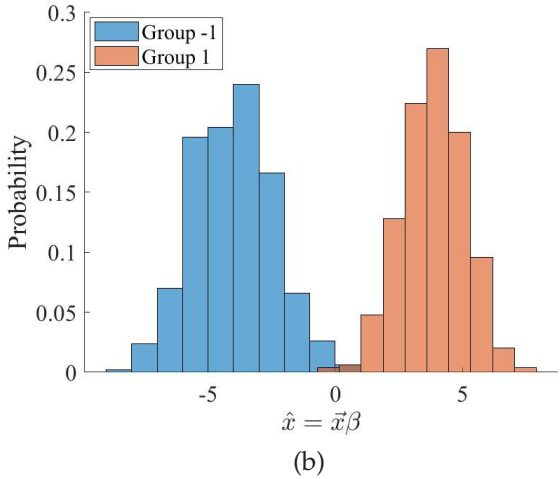}\\ 
	(b) 
	\end{tabular}
	\centering	
	\caption{The distribution of the instances before (a) and after (b) optimization for $ \beta $. Based on the optimized transformation instances become separable by $ \mathcal{D} $ in Eq. \ref{Eq:distinguisherLCC}.}
		\label{fig:LCC-example-hist}
\end{figure}

The histogram of the transformed instances before and after optimizing $ \beta $, shown in Fig. \ref{fig:LCC-example-hist}, indicates increased separability of instances. 

\subsection{Alternative discriminators} \label{sec:alternativediscriminators}
The most basic and intuitive discriminator for LCC, introduced in Eq. \ref{Eq:distinguisherLCC}, is based on the assumption that the instances are actually closer to their own center than to the other center in the transformed space. In this case, the classification procedure is straightforward: collect the training dataset, optimize $ \beta $ by solving the linear program in Eq. \ref{Eq:LCC}, and finally use Eq. \ref{Eq:distinguisherLCC} for discrimination. To classify a new instance $ \vec{x}_0 $ simply calculate $ \mathcal{M}_\beta(\vec{x}_0) $ and then use Eq. \ref{Eq:distinguisherLCC} for assigning the class label.

Since instances are centralized through the optimal choice of $ \beta $, the transformed instances in each class become closer to other instances from their class. This observation promotes the use of the 1-nearest neighbor (1-NN) \cite{altman1992introduction} classifier where the classification procedure becomes: collect the training dataset, optimize $ \beta $ by solving the linear program in Eq. \ref{Eq:LCC}, transform the training instances by $ \mathcal{M}_\beta $, and train 1-NN on the transformed data. To classify a new instance $ \vec{x}_0 $, first calculate $ \mathcal{M}_\beta(\vec{x}_0) $ and then use the trained 1-NN for assigning the class label. We refer to this modification as LCC-1NN.

There is no guarantee that after transformation, instances in each class have a similar level of spread. Hence, the choice of $ \frac{\hat{C}_{-1}+\hat{C}_{1}}{2} $ for distinguishing classes can be improved. The best threshold should be the one that has the maximum marginal distance to the instances in both classes in the transformed space, as is the case in SVM. Accordingly, we propose the following procedure for classification: collect the training dataset, optimize $ \beta $ by solving the linear program in Eq. \ref{Eq:LCC}, transform the training instances by $ \mathcal{M}_\beta $, and train a 1-dimensional SVM problem (note, the transformed instances are 1-dimensional) on the transformed data. For classification of a new instance $ \vec{x}_0 $, first calculate $ \mathcal{M}_\beta(\vec{x}_0) $ and then use the trained SVM for assigning the class label. We refer to this method as LCC-1SV. After finding $ \beta $, the center of the instances in the transformed space might be very close to one another, which may make SVM inefficient with an arbitrary value of $ \lambda $ (i.e., constraint satisfaction is ignored completely). Hence, after application of the transformation $ \mathcal{M}_\beta $ we scale the instances by $ s=\frac{h}{\hat{C}_{1}-\hat{C}_{-1}} $, where $ h $ is a constant (set to 10 in all of our experiments). This approach ensures that the instances from different classes are further away from one another by $ h $ units on average. In turn, allowing SVM to effectively distinguish between the 1-D instances using $\lambda=1$.

\subsection{Kernelization for LCC}
We provide an alternative representation based on dot product for any arbitrary instance $ \vec{z} $, $ \hat{C}_{-1} $, $ \hat{C}_{1} $, and $ \hat{l} $ in Eq. \ref{Eq:LCC}. This representation then enables the use of kernel trick in LCC to classify non-linearly separable instances.

Using the representer theorem \cite{scholkopf2001generalized}, the transformation $\beta$ in LCC can be expressed as:
\begin{equation}
\beta=\sum_{i=1}^{m}\alpha_i\vec{x}^{(i)}
\end{equation}
\noindent where $ \alpha_i \in \mathbb{R} $. The alternative representation of an arbitrary instance $ \vec{z} $ transformed by $ \beta $ is:
\begin{equation}
\hat{z}=\vec{z}\beta=\sum_{i=1}^{m}\alpha_i\vec{z}\vec{x}^{(i)}
\end{equation}
\noindent that requires calculation of $ \vec{z}\vec{x}^{(i)} $. For $ \hat{C}_{c} $, $ c \in \{-1,1\} $, we have:
\begin{equation}
\hat{C}_{c}=\frac{1}{m_c}\sum_{j}\vec{x}^{(j)}  \sum_{i=1}^{m}\alpha_i\vec{x}^{(i)}=\frac{\sum_{j}\sum_{i=1}^{m}\alpha_i\vec{x}^{(j)}\vec{x}^{(i)}}{m_c}
\end{equation}
\noindent where $ j\in\{j|y^{(j)}=c\} $, $ c $ is the class label ($ -1 $ or $ 1 $), $ m_k $ is the number of instances in the class $ c $, and $ \hat{C}_{c}=\vec{C}_{c}\beta $. Trivially, $ \hat{l} $ is calculated by $ \frac{\hat{C}_{-1}+\hat{C}_{1}}{2} $ in any space. By using these alternative representations, the optimization problem for LCC (Eq. \ref{Eq:LCC}) makes use of dot products in reference to instances. Hence, the kernel trick discussed in Section \ref{sec:kernalization} can be applied to handle non-linear classification problems . 

%% file: experiments.tex
\subsection{Outline of comparisons}
Here we introduce the datasets, pre-processes, and algorithm specific settings used in the comparisons. 
\subsubsection{Datasets}
We use eight datasets to compare classifiers, namely, Breast cancer (BC) , Crab gender (CG), Glass chemical (GC), Parkinson (PR), Ionosphere (IS), Pima Indians diabetes (PF), German credit card (GR),\footnote{All of these datasets are available online at https://archive.ics.uci.edu/ml/datasets.html} and Seizure detection (SD) \cite{michalski1986multi,temko2015detection,smith1988using,sigillito1989classification,parkinsonDb}. The main characteristics of these datasets are provided in Table \ref{Tab:dbs}. These datasets are used frequently as standard benchmarks in machine learning studies. 
		
\begin{table}[]
	\centering
	\caption{The datasets used for comparison purposes in this paper. $ n $ is the number of variables and $ c $ is the number of classes in each dataset. The number of instances in each class has been reported in the last column.}
	\label{Tab:dbs}
	\begin{tabular}{|l|l|l|l|}
		\hline
		\textbf{Dataset name} & \textbf{$ n $} & \textbf{$ c $} & \begin{tabular}[c]{@{}l@{}}\textbf{\# instance}\\ \textbf{in each class}\end{tabular}\\ \hline
		BC & 9 & 2 & $ <458,241> $ \\ \hline
		CG & 6 & 2 & $ <100,100> $ \\ \hline
		GC & 9 & 2 & $ <51,163> $ \\ \hline
		PR & 22 & 2 & $ <48,147> $ \\ \hline
		IS & 32 & 2 & $ <225,126> $ \\ \hline
		PD & 8 & 2 & $ <268,500> $ \\ \hline
		GR & 24 & 2 & $ <300,700> $ \\ \hline			
		SD* & ? & 2 & ? \\ \hline
	\end{tabular}
	\vspace{1ex}
	
	\raggedright *The seizure detection dataset includes 12 patients, each of them has their own number of variables and instances in different classes. See Table \ref{Tab:SDDetails}.
\end{table}
	
The SD dataset includes interacranial electroencephalogram (iEEG) from 12 subjects (four dogs and eight humans) with variable number of channels (Table \ref{Tab:SDDetails} shows the details of this dataset)\cite{temko2015detection}. There are two classes in this dataset, namely seizure (ictal) and no-seizure (interictal), with variable number of instances and iEEG channels for each subject. While each seizure event may take up to 60 seconds, each instance labeled as ictal or interictal in the dataset consists of 1 second of an event from all iEEG channels. As the properties of the signals belonging to the same ictal event are likely to be similar, the inclusion of different segments of a single event in both training and test sets may simplify the problem. Thus, we used all ictal segments that belonged to the same seizure event in either test or training set, but not both. The segment index associated with this dataset was used to reconstruct the events. This procedure is usually used for cross-validation in the seizure detection and prediction literature \cite{temko2015detection}. 
	
\begin{table}[]
	\centering
	\caption{Details of the seizure detection (SD) dataset. The first value in the last column is the number of instances of seizure (one second each) and the second number is the number of instances of non-seizure (one second each). The seizure instances are one second segments from different seizure events.}
	\label{Tab:SDDetails}
	\begin{tabular}{|l|l|l|l|}
		\hline
		\textbf{Patient} & \textbf{\# of Channels} & \begin{tabular}{c} \textbf{Sampling}\\\textbf{ rate (Hz)} 
		\end{tabular}& \begin{tabular}[c]{@{}l@{}}\textbf{\# instance}\\ \textbf{in each class}\end{tabular}\\ \hline
		Subject 1 & 16 & 400 & $ <178,418> $ \\ \hline
		Subject 2 & 16 & 400 & $ <172,1148> $ \\ \hline
		Subject 3 & 16 & 400 &  $ <480,4760> $ \\ \hline
		Subject 4 & 16 & 400 &  $ <257,2790> $ \\ \hline
		Subject 5 & 68 & 500 &  $ <70,104> $ \\ \hline
		Subject 6 & 16 & 5000 &  $ <151,2990> $ \\ \hline
		Subject 7 & 55 & 5000 &  $ <327,714> $ \\ \hline
		Subject 8 & 72 & 5000 &  $ <20,190> $ \\ \hline
		Subject 9 & 64 & 5000 &  $ <135,2610> $ \\ \hline
		Subject 10 & 30 & 5000 &  $ <225,2772> $ \\ \hline
		Subject 11 & 36 & 5000 &  $ <282,3239> $ \\ \hline
		Subject 12 & 16 & 5000 &  $ <180,1710> $ \\ \hline
	\end{tabular}
\end{table}
	
\subsubsection{Preprocessing and performance measures}	We preprocessed the instances in the SD dataset by calculating the fast Fourier transform of each channel and concatenated transformed signals to generate one large signal (FFT of the channels one after another). The length of this signal is a function of the number iEEG channels. We used frequencies from 1 to 50 Hz only as this shows a sufficiently accurate presentation of a seizure \cite{temko2015detection}. For subject 1, for example, the preprocessed signal was $ 16 \times 49 $ samples long (16 channels, 1 Hz to 50 Hz FFT). It is noticeable that the number of instances in each class for the SD dataset is imbalanced (the ratio of ictal to interictal instances is about 2:19 on average). The data from each patient was considered as a unique independent dataset. The other datasets were used in their original form without specific preprocessing. Two performance evaluation procedures were considered. 

\textbf{Performance evaluation procedure 1)} in each run, a training set was generated by randomly selecting 70\% of instances in each class, i.e., stratified sampling rather than pure random sampling. This sampling method has been shown \cite{kohavi1995study} to be more effective than pure random sampling. The reminder of the instances in each class were used for testing. All methods (LDA, SVM, WSV, TSV, SMP, DEL, DLS, and LCC) were then trained on the training set and evaluated on the test set. The final results reported in this paper for this procedure are based on 100 independent runs, each run the training and the test sets were remained the same for all methods. SMP was excluded from all tests that involved the SD dataset as it took longer than 300 seconds (maximum training time in our tests) to train for each subject because of the large number of variables. We used the Wilcoxon rank test (confidence of 0.05) for statistical comparisons between LCC and other methods. The default parameter values were used for all methods. 

\textbf{Performance evaluation procedure 2)} for each dataset (except SD) and method, we conducted a 10-fold stratified cross validation and picked the parameter values for which the average performance of the method on the test fold (the left-out fold) was maximized. 

The variables in the training set in both evaluation procedures were normalized to have a mean of zero and standard deviation of 1. The mapping used to normalize the training set was applied to the test set to ensure that both training and test sets were in the same domain. The variables with zero variance across the entire dataset were removed as some of the tested methods had difficulty to deal with such variables. Area Under the Curve (AUC) of the Receiver Operating Characteristic (ROC) \cite{hanley1982meaning} was used as a performance measure. 

\subsubsection{Algorithm settings}
We compared the performance of LCC with SVM, SMP, TSV, DLS, DEL, WSV, and LDA. We used MATLAB 2017a for implementations and tests.\footnote{The source code for LCC is available as a supplementary to this article.} 

For our performance evaluation procedure (1) we used recommended values obtained from related papers, while we used the optimized parameter values, i.e. obtained by searching within a range of values, for our performance evaluation procedure (2): 
\begin{itemize}
	\item DEL: 400 iterations, $ \lambda=.5 $, $ \mu=1e-8 $. The tested values for cross validation of $ \lambda $ in the second performance evaluation procedure were $ \{1/15, 2/15,...,1\} $.
	\item TSV: $c_1=0.1$, $c_2=0.1$, $c_3=0.1$, $c_4=0.1$, as suggested in the source code. The tested values for cross validation in the second performance evaluation procedure were $ \{10^{-30/15},10^{-26/15},...,10^{26/15}\} $.
	\item WSV: $c_1=0.1$, $c_2=0.1$, $c_3=0.1$, $c_4=0.1$, as suggested in the source code. The tested values for cross validation in the second performance evaluation procedure were $ \{10^{-30/15},10^{-26/15},...,10^{26/15}\} $.
	\item SMP: none (no parameters to be optimized)
	\item DLS: $ \lambda = 0.5 $. The tested values for cross validation in the second performance evaluation procedure were $ \{1/15, 2/15,...,1\} $.	
	\item SVM: $ \lambda=1 $. The tested values for cross validation in the second performance evaluation procedure were $ \{10^{-30/15},10^{-26/15},...,10^{26/15}\} $.
	\item LDA: $ \lambda = 0.5 $. The tested values for cross validation in the second performance evaluation procedure were $ \{1/15, 2/15,...,1\} $.	
    \item LCC: $ \lambda = 2 $, $\sigma = -0.01$. The tested values for cross validation of $ \sigma $ in the second performance evaluation procedure were $ \{-2^{-7},-2^{-6},...,-2^{7}\} $.
\end{itemize}

Ranges were selected in a way that only 15 distinct values were tested for each parameter.

\subsection{Numerical comparison between different centralization-based classifiers and discriminators}
\begin{figure}
	\centering
	\includegraphics[width=0.49\textwidth]{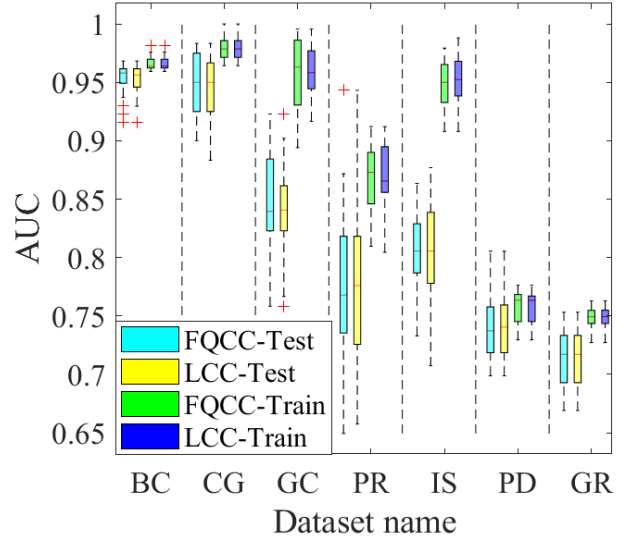} 
	\caption{Comparative results between FQCC and LCC.}
	\label{fig:Fqcc-vs-Lcc}
\end{figure}

We first compared FQCC to LCC to test whether our simplifications led to differences in results. Comparison results were generated from the standard benchmark problems listed in Table \ref{Tab:dbs} (see Fig. \ref{fig:Fqcc-vs-Lcc}). Figure \ref{fig:Fqcc-vs-Lcc} shows that the performance of FQCC and LCC in both training and testing datasets were almost the same. This was also confirmed by a Wilcoxon test where the performance of FQCC and LCC on training and testing datasets for all benchmark problems showed no significant difference ($ p \gg 0.05 $ in all cases). Notably, the average execution time for FQCC was almost 250 times longer than for LCC.

\begin{figure}
	\centering
	\includegraphics[width=0.49\textwidth]{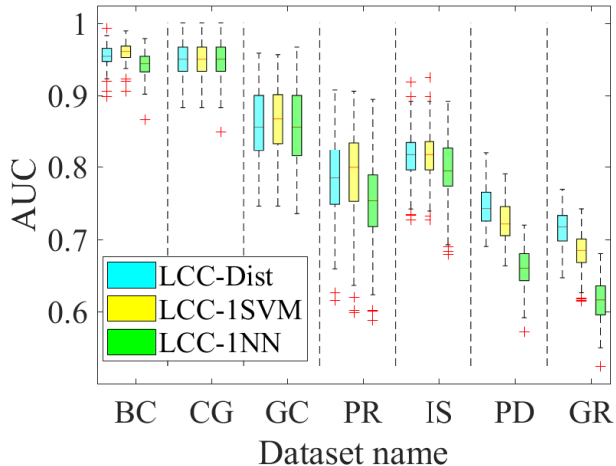} 
	\caption{Comparative results between FQCC and LCC using the distance discriminator (LCC-Dist, Eq. \ref{Eq:distinguisherLCC}), SVM-based (LCC-1SVM) and 1-nearest neighbour (LCC-1NN).}
    \label{fig:discriminator-test}
\end{figure}

In Fig. \ref{fig:discriminator-test} we compare the discriminators introduced in Section \ref{sec:alternativediscriminators}. The results show that the performance of LCC is maximized with the use of the LCC-Dist discriminator defined by Eq. \ref{Eq:distinguisherLCC}. Hence, we use LCC-Dist (LCC for short) in the rest of our experiments.\footnote{Note that this observation is only specific to this dataset and for any other dataset a cross validation procedure between these discriminators is recommended.} 

\subsection{Comparison with existing classifiers: default parameters}
\label{sec:experimentOverall}
Table \ref{Tbl:resultsCompact0} provides comparative results\footnote{Details of this experiment is available in Appendix A.} for methods tested across the datasets considered (top three rows: results for all datasets except SD, bottom three rows: results for SD dataset over 12 subjects). The value in the row labeled "Train" under each column indicates the number of datasets for which LCC significantly outperforms (Wilcoxon test, $ p < 0.05 $) other algorithms listed in the column. The values in the parentheses indicates the number of datasets for which the algorithm in that particular column significantly outpeformed LCC. For example, the entry $ 5(1) $ in the row labeled "Train" (top three rows) and column labeled "SMP", indicates that LCC outperformed SMP in 5 (out of 7) datasets while SMP performed significantly better than LCC once (the value of 1 inside the parentheses). 

\textbf{Top three rows} of the table show that for training, LCC performs significantly better than the tested methods in most datasets. In terms of testing, however, LCC competes closely with SMP and DEL, two state-of-the-art classification methods. LCC was found to be faster than SVM and SMP, while having similar execution time to DEL. LCC is significantly slower than LDA, WSV, TSV and DLS. 
	
\begin{table}[]
	\centering
	\caption{Comparison results between LCC and other classifiers when they were applied to the datasets in Table \ref{Tab:dbs}. \textbf{Top 3 rows:} Results for all datasets excluding SD. \textbf{Bottom 3 rows:} results for SD.}
	\label{Tbl:resultsCompact0}
	\begin{tabular}{|l|l|l|l|l|l|l|l|}
		\hline
		& \textbf{LDA} & \textbf{SVM} & \textbf{DLS} & \textbf{SMP} & \textbf{WSV} & \textbf{TSV} & \textbf{DEL} \\ \hline
		\textbf{Train} & 7(0) & 6(1) & N/A & 5(1) & 6(1) & 7(0) & 5(1) \\ \hline
		\textbf{Test} & 6(1) & 3(2) & 3(2) & 3(2) & 3(3) & 4(1) & 3(3) \\ \hline
		\textbf{Time} & 0(7) & 7(0) & 0(3) & 7(0) & 0(7) & 1(5) & 3(3) \\ \hline
		\textbf{Train} & 5(0) & 0(5) & 0(5) & - & 2(3) & 0(3) & 11(0) \\ \hline
		\textbf{Test} & 6(3) & 8(1) & 9(0) & - & 8(3) & 7(2) & 10(2) \\ \hline
		\textbf{Time} & 1(11) & 7(4) & 9(3) & - & 2(10) & 2(10) & 11(1) \\ \hline	
	\end{tabular}
\end{table}	

\textbf{Bottom three rows} of Table \ref{Tbl:resultsCompact0} indicate that LCC has performed significantly better than all other methods in the test datasets. As suggested by the results, for the training datasets LCC appears to only work better than LDA and DEL.

\subsection{Comparison with existing classifiers: optimized parameters}
To reduce the impact of the parameters on the performance of the methods, we perform a 10-fold stratified cross validation to set individual method parameters after which a comparison of methods is performed. The best performances are then ranked (0 indexed), corrected for rank repetitions, and presented in Table \ref{Tbl:results10Fold}. 

The value of $ \lambda $ for LCC was set to 2 as our experiments showed that this value works equally well for all datasets considered herein. Table \ref{Tbl:results10Fold} indicates that the average rank for LCC over all seven datasets, having optimized method parameters, is lower than rankings of other methods. This provides evidence that LCC performs better than the other tested methods on average across datasets considered.  

\begin{table}[]
	\centering
	\caption{10 fold cross validation with optimized parameters}
	\label{Tbl:results10Fold}
	\begin{tabular}{|l|l|l|l|l|l|l|l|}
		\hline
		\textbf{LCC} & \textbf{LDA} & \textbf{SVM} & \textbf{DLS} & \textbf{SMP} & \textbf{WSV} & \textbf{TSV} & \textbf{DEL} \\ \hline
		2.57 & 5 & 3.14 & 3.71 & 2.86 & 3.64 & 3.78 & 3.28 \\ \hline			
	\end{tabular}
\end{table}	

\subsection{Kernel test}
\begin{figure}
	\begin{tabular}{cc}
	\includegraphics[width=0.2\textwidth]{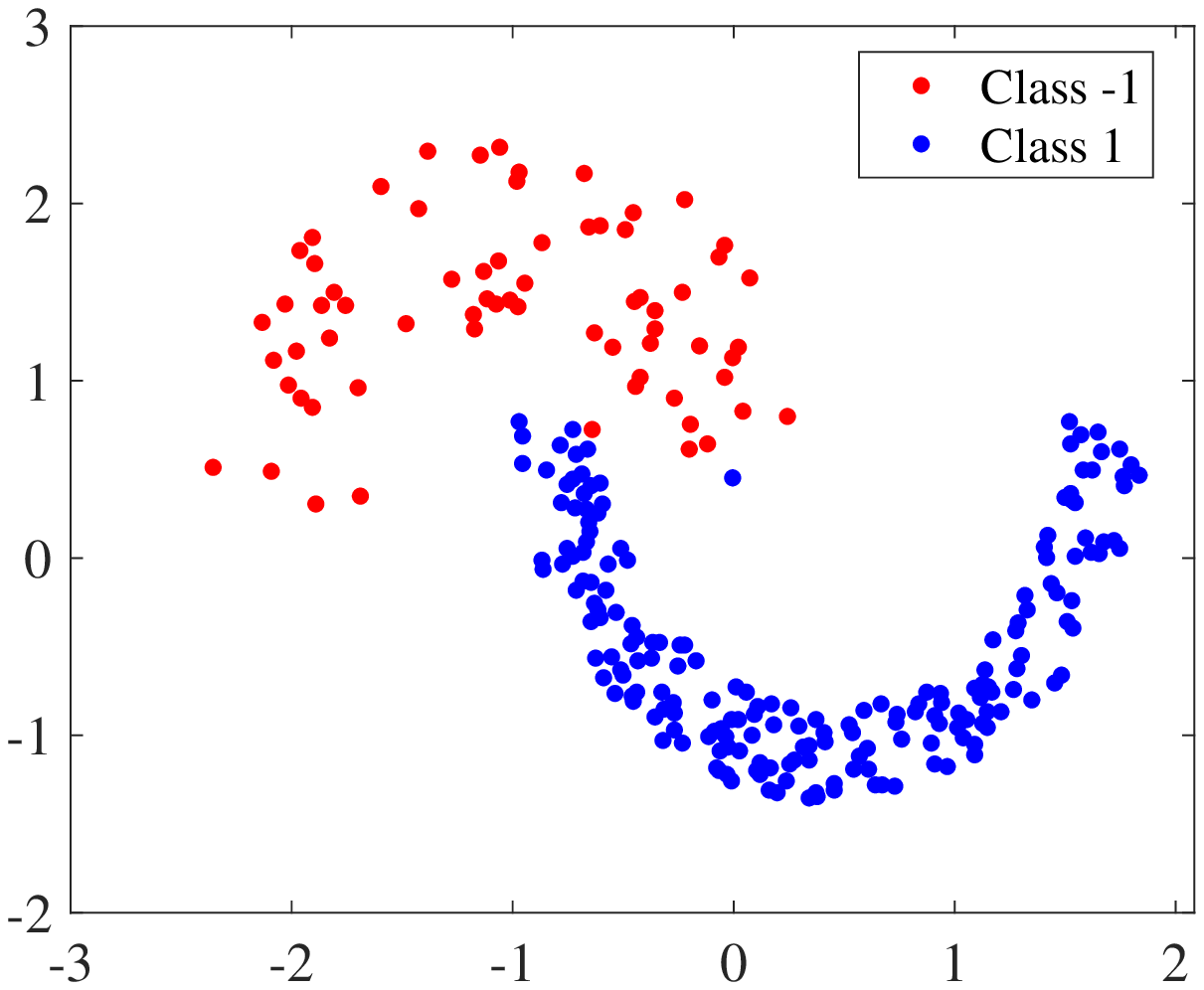} & \includegraphics[width=0.2\textwidth]{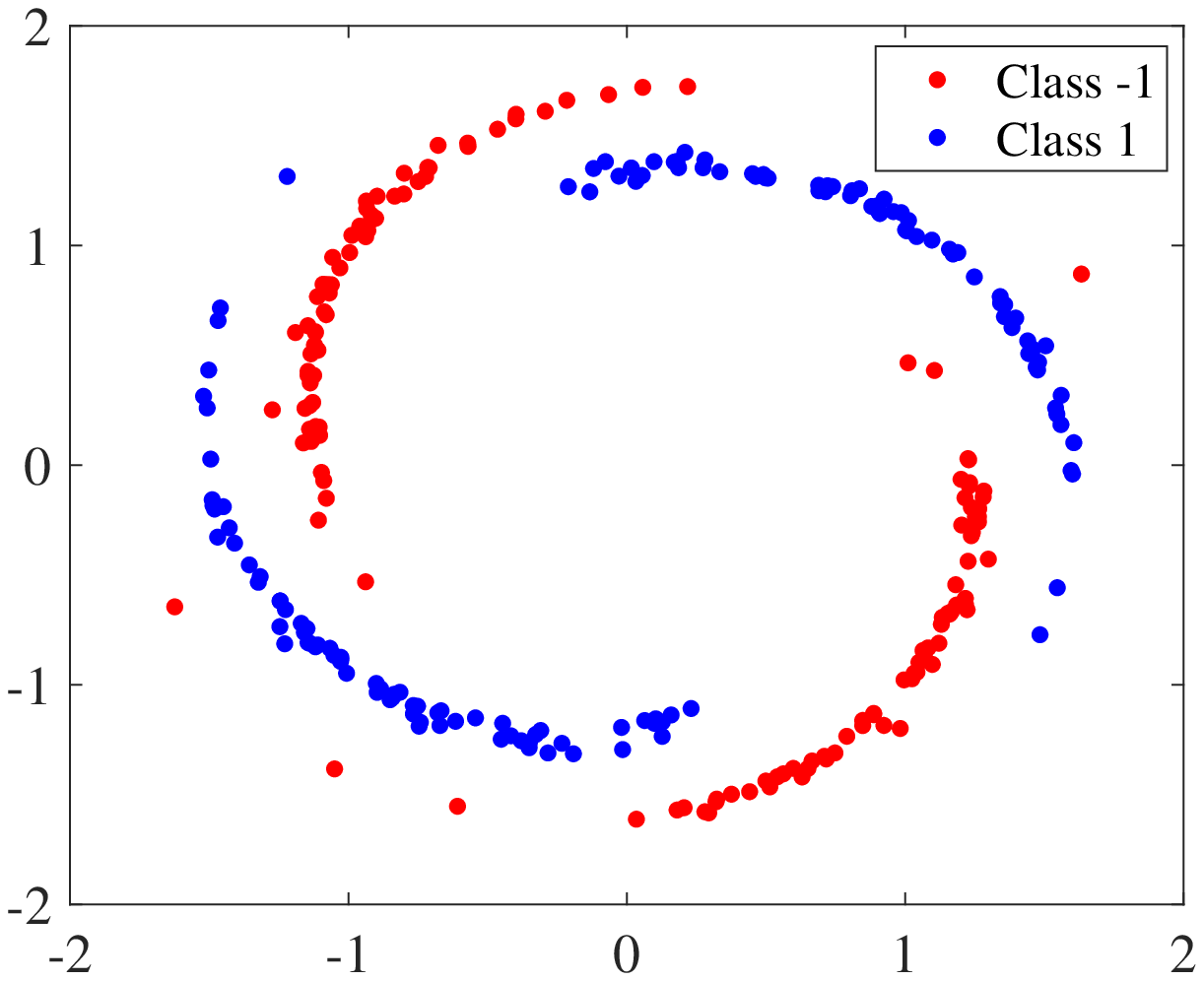}\\
	(a) & (b)\\		
	\includegraphics[width=0.2\textwidth]{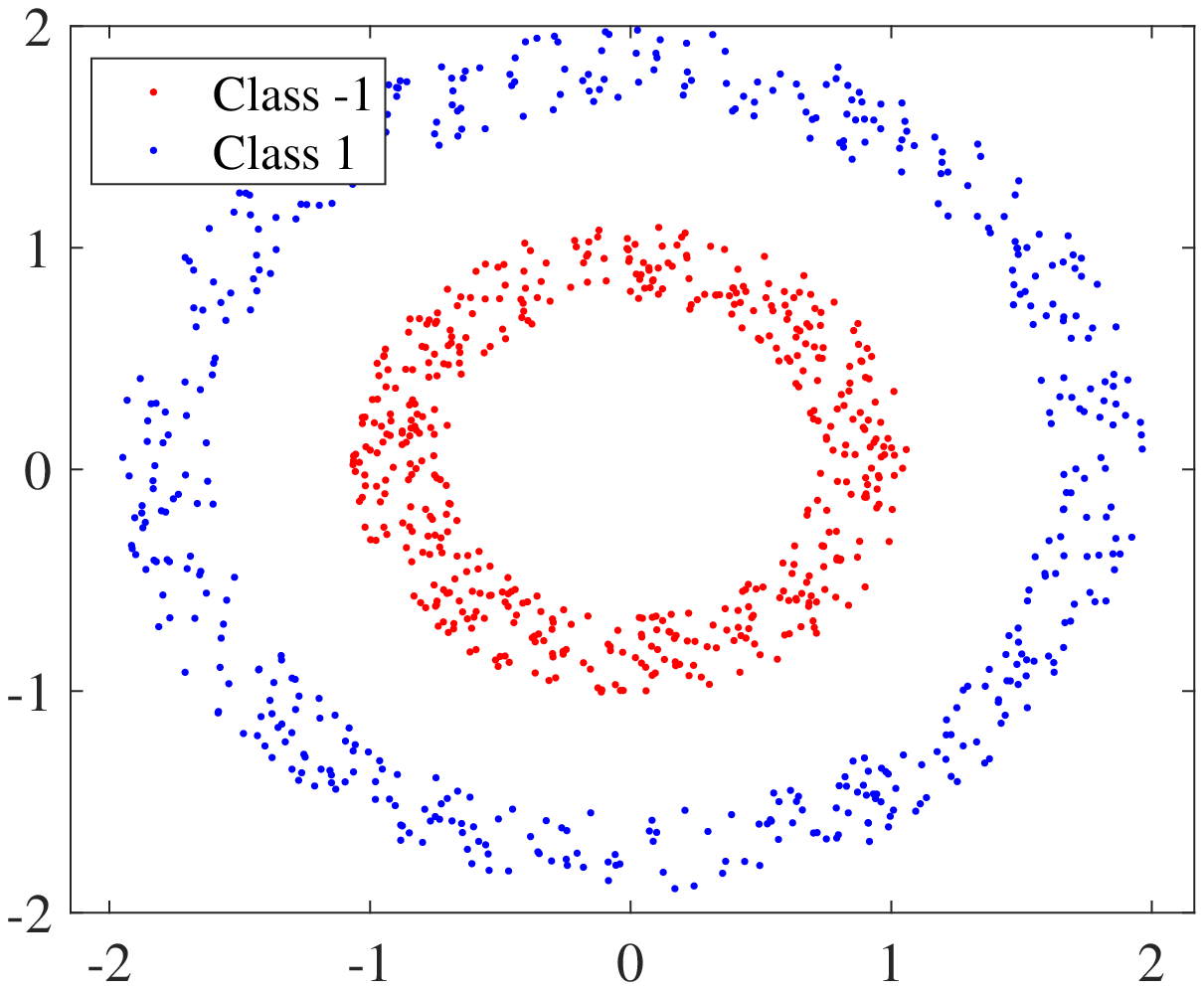} & \includegraphics[width=0.2\textwidth]{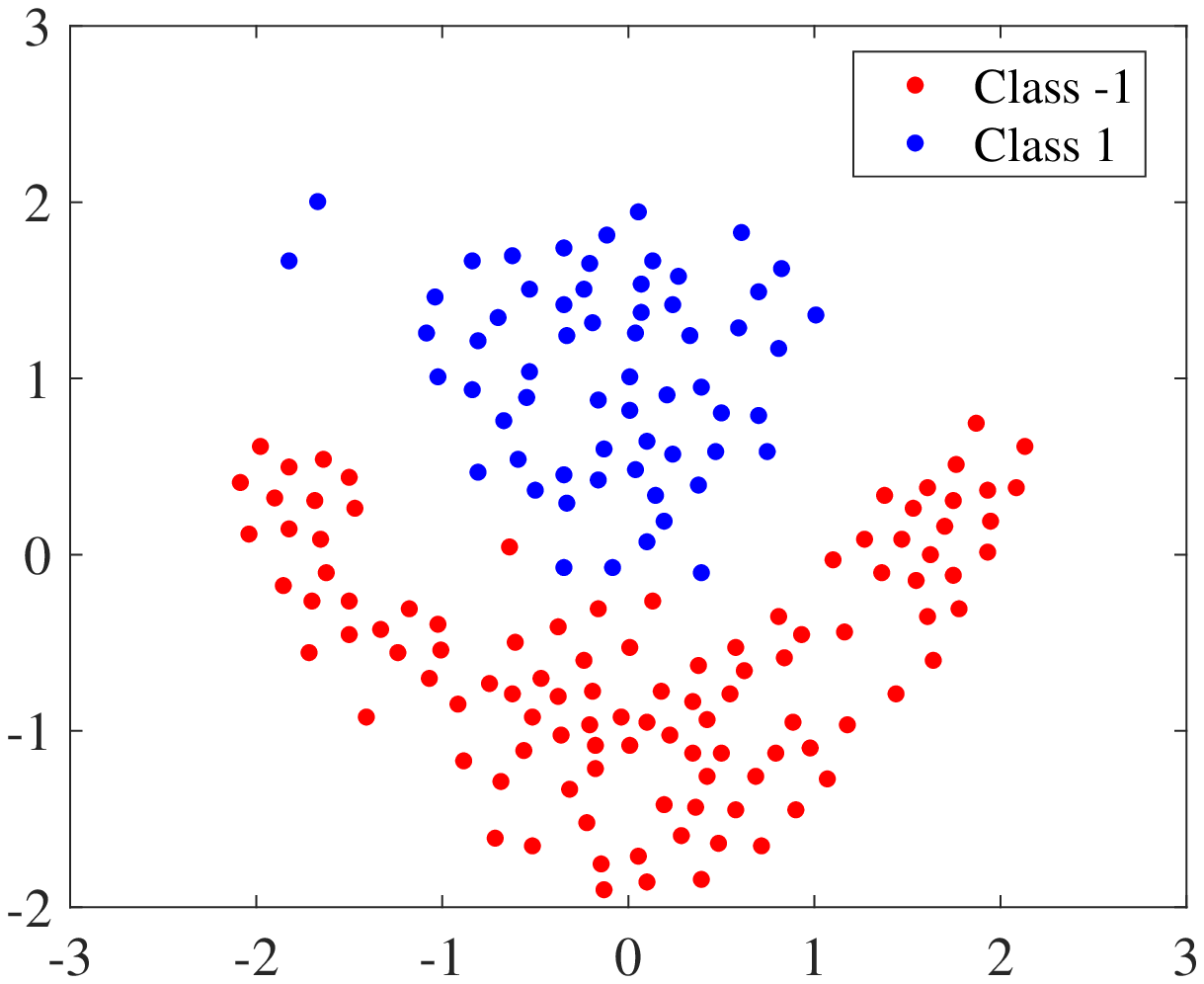}\\ 
	(c) & (d) 
	\end{tabular}
	\centering	
	\caption{Results for classification by kernelized (RBF) LCC: (a) Jain, (b) Spiral, (c) Circles, and (d) Flame.}
		\label{fig:LCC-kernel}
\end{figure}

We used four known synthetic datasets to test the performance of LCC with a kernel. We opted to use the RBF kernel for this test. The datasets include Jain \cite{jain2005data} (Figure \ref{fig:LCC-kernel}-a), spiral \cite{chang2008robust} (Figure \ref{fig:LCC-kernel}-b), circles \cite{chang2008robust} (Figure \ref{fig:LCC-kernel}-c), and flame \cite{Fu2007} (Figure \ref{fig:LCC-kernel}-d). In all of these datasets, the performance of LCC with RBF kernel was 100\% for both train and test except for the Jain dataset that the training and testing performances were 99.5\% and 98.2\% in average, respectively.

%% file: appendix.tex
\section*{Appendix-A}
Table \ref{Tbl:comparisonResultsDisc} shows the average results of 50 independent runs for all algorithms. The values in the table have been prefixed by a character to indicate the results of the statistical test (Wilcoxon test with confidence 0.05) between the method indicated in that column and LCC, for the dataset at that row.  "*", "-", and "+" indicate that the result is statistically worst, the same, or better than LCC. For example, the value "$95.13^{*}$" in the row "BC", measure "Train", column TSVM indicates that the average performance (AUC) of the method TSVM was $95.13$ on the training set, that was significantly worse than LCC.
		
\begin{table*}[]
	\centering
	\caption{Comparison of results of 6 well-known classification methods with LCC. The values in each row are the average values (over 50 runs). The row "Time" denotes the average time in milliseconds over 50 runs.}
	\label{Tbl:comparisonResultsDisc2}
	\begin{tabular}{|l|l|l|l|l|l|l|l|l|l|}
		\hline
		Dataset & Measure & \textbf{DEL} & \textbf{TSV} & \textbf{WSV} & \textbf{SMP} & \textbf{DSL} & \textbf{SVM} & \textbf{LDA} & \textbf{LCC} \\ \hline
\multirow{4}{*}{BC}  & Time & 30.1+ & 59.4* & 0.9+ & 197.6* & 17.8+ & 90.2* & 10.7+ & 26.5  \\ \cline{2-10} 
& Train & 96.95+ & 95.18* & 94.52* & 97.48+ & \textbf{100}+ & 96.9+ & 95.12* & 96.42  \\ \cline{2-10} 
& Test & 96.61+ & 94.88* & 94.2* & \textbf{97.26}+ & 94.88* & 96.31+ & 94.92* & 95.58  \\ \hline
\multirow{4}{*}{CG}  & Time & 20.1* & 4.1+ & 0.4+ & 177.2* & 14.6- & 31.3* & 7.3+ & 15.2  \\ \cline{2-10} 
& Train & 94.12* & 95.89* & 96.74* & 95.64* & \textbf{100}+ & 95.4* & 92.11* & 98.09  \\ \cline{2-10} 
& Test & 94.12* & 95.4- & \textbf{95.75}- & 94.73* & 94.97- & 94.7* & 89.65* & 95.43  \\ \hline
\multirow{4}{*}{GC}  & Time & 10.8- & 6+ & 0.4+ & 168* & 11.2- & 28* & 6.2+ & 13.7  \\ \cline{2-10} 
& Train & 92.29* & 88.94* & 88.51* & 93.43* & \textbf{100}+ & 91.9* & 88.52* & 94.26  \\ \cline{2-10} 
& Test & 89.31+ & 85.51* & 85.41* & \textbf{90.67}+ & 87.93+ & 87.21- & 85.73* & 87.03  \\ \hline
\multirow{4}{*}{PR}  & Time & 20.8* & 3+ & 0.6+ & 231* & 17.5- & 32.3* & 7.5+ & 17.7  \\ \cline{2-10} 
& Train & 79.54* & 84.95* & 88.05+ & 85.74* & \textbf{100}+ & 82.73* & 75.65* & 86.94  \\ \cline{2-10} 
& Test & 76.31* & 78.47- & \textbf{80.45}+ & 76.61* & 78.28- & 77.38- & 74.59* & 78.35  \\ \hline
\multirow{4}{*}{IS}  & Time & 30.2+ & 5.8+ & 0.8+ & 234.1* & 21.4- & 36.1* & 7.6+ & 21.8  \\ \cline{2-10} 
& Train & 87.34* & 85.95* & 70.67* & 89.68* & \textbf{100}+ & 94.49* & 86.67* & 94.95  \\ \cline{2-10} 
& Test & 81.16- & 79.49* & 67.38* & 82.98+ & 82.64+ & \textbf{83.97}+ & 83.95+ & 81.17  \\ \hline
\multirow{4}{*}{PD}  & Time & 20.7+ & 10.8+ & 0.7+ & 174.8* & 7.1+ & 81.7* & 7.5+ & 37  \\ \cline{2-10} 
& Train & 75.35- & 74.52* & 74.55* & 75.53- & \textbf{100}+ & 73.15* & 72.54* & 75.53  \\ \cline{2-10} 
& Test & \textbf{75.04}+ & 73.81* & 73.88* & 74.77- & 65.53* & 72.38* & 71.69* & 74.57  \\ \hline
\multirow{4}{*}{GR}  & Time & 40.8* & 30.4- & 0.9+ & 161.8* & 6.9+ & 116.9* & 7.4+ & 30.5  \\ \cline{2-10} 
& Train & 73.3* & 74.34* & 74.24* & 74.35* & \textbf{100}+ & 71.28* & 71.41* & 74.68  \\ \cline{2-10} 
& Test & 70.98* & \textbf{71.88}+ & 71.75- & 70.86* & 62.07* & 68.28* & 68.97* & 71.57  \\ \hline
	\end{tabular}
\end{table*}
	
Table \ref{Tbl:comparisonResultsDisc} shows the average results of 50 independent runs for all algorithms when they were applied to SD. The values in the table have been postfixed by a character, defined as in Table \ref{Tbl:comparisonResultsDisc2}.
	
	\begin{table*}[]
		\centering
		\caption{Comparison of results of 6 well-known classification methods with LCC. The values in each row are the average values (over 50 runs). The row "Time" denotes the average time in milliseconds over 50 runs.}
		\label{Tbl:comparisonResultsDisc}
		\begin{tabular}{|l|l|l|l|l|l|l|l|l|}
			\hline
			Dataset & Measure & \textbf{DEL} & \textbf{TSV} & \textbf{WSV} & \textbf{DSL} & \textbf{SVM} & \textbf{LDA} & \textbf{LCC} \\ \hline
\multirow{4}{*}{Subject1}  & Time & 3860.9* & 48.4+ & 58.6+ & 2189.1* & 896.3* & 239+ & 307.1  \\ \cline{2-9} 
& Train & 85.19* & \textbf{100}+ & 100+ & 100+ & 100+ & 99.6- & 99.32  \\ \cline{2-9} 
& Test & 80.28* & 92.04* & 93.61* & 92.22* & 95.15* & 94.97* & \textbf{95.2}  \\ \hline
\multirow{4}{*}{Subject2}  & Time & 4496.4* & 101.6+ & 83.8+ & 4132.7* & 410.3* & 195.6+ & 349.3  \\ \cline{2-9} 
& Train & 88.51* & \textbf{100}+ & 100+ & 100+ & 100+ & 98.55* & 99.13  \\ \cline{2-9} 
& Test & 70.72* & 80.73+ & \textbf{84.2}+ & 77.2- & 79.44- & 80.51+ & 77.01  \\ \hline
\multirow{4}{*}{Subject3}  & Time & 8561.3* & 1753+ & 232.1+ & 14754.1* & 10690.1* & 471.8+ & 2879.4  \\ \cline{2-9} 
& Train & 91.16* & 96.5- & 96.44* & \textbf{100}+ & 100+ & 95.07* & 96.67  \\ \cline{2-9} 
& Test & 85.24* & 91.54* & 91.75* & 88.46* & 89.44* & 91.47* & \textbf{92.51}  \\ \hline
\multirow{4}{*}{Subject4}  & Time & 6467.3* & 342.3+ & 149+ & 7948.1* & 3015+ & 307.3+ & 4447.2  \\ \cline{2-9} 
& Train & 82.39* & 98.08+ & 96.77+ & \textbf{100}+ & 100+ & 92.62- & 91.42  \\ \cline{2-9} 
& Test & 75.44* & 82.5* & 86.88* & 74.11* & 77.13* & 94.37* & \textbf{97.89}  \\ \hline
\multirow{4}{*}{Subject5}  & Time & 86772.1* & 509.1* & 583.7* & 47.6+ & 126.5+ & 149.6+ & 258.4  \\ \cline{2-9} 
& Train & \textbf{100}- & 100- & 100- & 100- & 100- & 100- & 100  \\ \cline{2-9} 
& Test & 73.31* & 85.62- & \textbf{89.84}+ & 77.09* & 84.09* & 80.46* & 85.42  \\ \hline
\multirow{4}{*}{Subject6}  & Time & 6580.6* & 382.4+ & 153.3+ & 10840.1* & 2536.8* & 321.2+ & 800.8  \\ \cline{2-9} 
& Train & 86.49* & 99.85- & 99.68- & \textbf{100}+ & 100+ & 97.38- & 97.5  \\ \cline{2-9} 
& Test & 84.93* & \textbf{97.86}- & 97.01- & 94.53* & 95.03* & 96.77* & 97.26  \\ \hline
\multirow{4}{*}{Subject7}  & Time & 57439.2* & 474.2+ & 636.8+ & 157.1+ & 381.4+ & 577.3+ & 1388.1  \\ \cline{2-9} 
& Train & 84.49* & \textbf{100}- & 100- & 100- & 100- & 100- & 100  \\ \cline{2-9} 
& Test & 65.27+ & \textbf{65.86}+ & 60.18+ & 59.83- & 61.44+ & 63.87+ & 58.63  \\ \hline
\multirow{4}{*}{Subject8}  & Time & 98915.1* & 572* & 664.9* & 24.8+ & 36+ & 123.5+ & 217.6  \\ \cline{2-9} 
& Train & 96.05* & \textbf{100}- & 100- & 100- & 100- & 100- & 100  \\ \cline{2-9} 
& Test & \textbf{62.95}+ & 55.07* & 52.79* & 57.82- & 53.75* & 56.2- & 56.23  \\ \hline
\multirow{4}{*}{Subject9}  & Time & 85342.2* & 1291+ & 1853.9+ & 58231* & 1935.1- & 7843* & 2365.7  \\ \cline{2-9} 
& Train & 92.21* & \textbf{100}- & 100- & 100- & 100- & 99.57* & 100  \\ \cline{2-9} 
& Test & 64.31* & 69.9* & 71.45* & 79.44* & 85.02- & \textbf{89.63}+ & 86.5  \\ \hline
\multirow{4}{*}{Subject10}  & Time & 19087* & 543+ & 439.4+ & 22046.1* & 2149.9* & 1105.1+ & 1350  \\ \cline{2-9} 
& Train & 88.47* & \textbf{100}- & 99.96- & 100- & 100- & 98.41* & 100  \\ \cline{2-9} 
& Test & 80.39* & 85.65* & 85.88* & 86.71* & 85.84* & 87.46* & \textbf{87.91}  \\ \hline
\multirow{4}{*}{Subject11}  & Time & 28347.4* & 796.6+ & 734.2+ & 30321.8* & 3051.2* & 1909.4+ & 2812  \\ \cline{2-9} 
& Train & 88.46* & \textbf{100}- & 99.21* & 100- & 100- & 95.37* & 100  \\ \cline{2-9} 
& Test & 91.31* & 98.05* & 97.89* & 98.13* & 99.03- & \textbf{99.37}- & 99.2  \\ \hline
\multirow{4}{*}{Subject12}  & Time & 5155.7* & 184.5+ & 99.6+ & 7578.5* & 717.2* & 248.9+ & 460.6  \\ \cline{2-9} 
& Train & 89.09* & 99.97- & 99.49- & \textbf{100}- & 100- & 97.3- & 96.62  \\ \cline{2-9} 
& Test & 80.27* & 94.74- & 93.73* & 88.27* & 91.77* & \textbf{95.18}- & 94.57  \\ \hline
		\end{tabular}
\end{table*}